\documentclass[]{fairmeta}
\usepackage{amsthm,amsmath,amssymb}
\usepackage{graphicx}
\usepackage{array}
\usepackage{natbib}
\usepackage{tcolorbox}
\usepackage{subcaption}
\definecolor{techblue}{RGB}{240, 248, 255}

\newtheorem{theorem}{Theorem}[]

\newtheorem{remark1}[theorem]{Remark}

\usepackage{xcolor}

\title{Tele-Omni: a Unified Multimodal Framework for Video Generation and Editing}
\author{TeleAI}
\metadata[Corresponding Author]{Xuelong Li(\email{xuelong$\_$li@ieee.org})}

\begin{document}

\abstract{

Recent advances in diffusion-based video generation have substantially improved visual fidelity and temporal coherence. However, most existing approaches remain task-specific and rely primarily on textual instructions, limiting their ability to handle multimodal inputs, contextual references, and diverse video generation and editing scenarios within a unified framework. Moreover, many video editing methods depend on carefully engineered pipelines tailored to individual operations, which hinders scalability and composability.
In this paper, we propose \textbf{Tele-Omni}, a unified multimodal framework for video generation and editing that follows multimodal instructions, including text, images, and reference videos, within a single model. Tele-Omni leverages pretrained multimodal large language models to parse heterogeneous instructions and infer structured generation or editing intents, while diffusion-based generators perform high-quality video synthesis conditioned on these structured signals. To enable joint training across heterogeneous video tasks, we introduce a task-aware data processing pipeline that unifies multimodal inputs into a structured instruction format while preserving task-specific constraints.
Tele-Omni supports a wide range of video-centric tasks, including text-to-video generation, image-to-video generation, first-last-frame video generation, in-context video generation, and in-context video editing. By decoupling instruction parsing from video synthesis and combining it with task-aware data design, Tele-Omni achieves flexible multimodal control while maintaining strong temporal coherence and visual consistency. Experimental results demonstrate that Tele-Omni achieves competitive performance across multiple tasks.

}

\maketitle

\vspace{-0.1em}

\section{Introduction}
Recent years have witnessed remarkable progress in artificial intelligence, driven by rapid advances in large language models and diffusion-based generative models~\citep{wei2025univideo,wu2025janus}. In the visual domain, diffusion models have demonstrated strong capabilities in video synthesis, achieving impressive visual fidelity and temporal coherence. In parallel, the emergence of multimodal large language models (MLLMs)~\citep{qwen3,bai2023qwen} has substantially enhanced the flexibility and expressiveness of instruction-driven systems. Together, these developments provide a solid foundation for building controllable and interactive frameworks for video generation and editing.

Despite this progress, existing video generation systems remain largely task-specific. Most diffusion-based video models~\citep{zheng2024open,singer2022make,Cogvideo,gupta2024photorealistic} are designed for a single task—most commonly text-to-video generation—and rely almost exclusively on textual prompts for control. Such designs impose clear practical limitations. On the one hand, text alone is often insufficient to precisely specify complex visual intent, and users frequently need to rely on reference images or example videos to convey appearance, motion patterns, or stylistic preferences. On the other hand, single-task formulations make it difficult for models to extend beyond basic generation, particularly to more complex scenarios such as context-aware generation or video editing.

Meanwhile, current video editing methods~\citep{liu2024video,tu2025videoanydoor,wang2024replace,minimaxremover} typically adopt task-specific pipelines tailored to individual operations, such as object replacement, attribute modification, or localized editing. Although effective in constrained settings, these approaches suffer from limited scalability and flexibility. Different editing operations often require distinct model architectures or processing workflows, making them difficult to integrate into a unified system. Moreover, such methods rarely support composing multiple editing operations within a single instruction or handling scenarios that depend on contextual references.

As a result, important capabilities—including multimodal instruction-driven video generation, in-context video generation, and flexible video editing—remain insufficiently addressed by existing approaches. Current solutions either focus on isolated generation tasks or rely on heavily engineered pipelines, limiting their applicability in realistic settings where multimodal inputs and diverse operations must be handled jointly.

To address these challenges, we propose Tele-Omni, a unified multimodal framework for video generation and video editing. Tele-Omni is designed to follow multimodal instructions, including text, images, and reference videos, within a single model, enabling a wide range of video-centric generation and editing tasks without relying on task-specific architectures or manually designed processing pipelines. The core goal of Tele-Omni is to improve adaptability to multimodal instructions and diverse video tasks while maintaining high visual quality and strong temporal coherence.

Building such a unified video framework presents challenges not only in model design but also in how training data is organized~\citep{jiang2025vace,wei2025univideo}. Different video generation and editing tasks vary substantially in their conditioning inputs, and target outputs. For example, generation tasks emphasize global content structure and motion plausibility, whereas editing tasks require modifying specific regions while preserving the remaining content~\citep{tan2024vidgen,villegas2022phenaki,openve,yuan2025opens2v,wang2025koala}. Naively mixing heterogeneous datasets across tasks often leads to unstable training and degraded performance.

To this end, we systematically design task-aware data processing pipelines that enable multiple video generation and editing tasks to be jointly trained within a unified framework. Specifically, Tele-Omni is trained on a diverse collection of video tasks, including text-to-video generation, image-to-video generation, first-last-frame video generation, in-context video generation, and in-context video editing. For each task, we construct a dedicated data organization strategy that unifies multimodal inputs into a structured instruction representation while explicitly encoding task-specific constraints.

In this formulation, textual prompts describe the intended generation or editing objective, reference images or videos provide appearance, motion, or temporal constraints, and target videos serve as supervision signals. This unified yet task-aware representation allows the model to infer the intended operation directly from multimodal instructions, without requiring explicit task labels or separate task-specific prediction heads.

The data pipelines are further designed with an emphasis on temporal consistency and visual coherence. For generation tasks, training samples encourage stable motion patterns and consistent appearance across frames. For editing tasks, original and edited videos are constructed in paired form, enabling the model to distinguish editable regions from content that should remain unchanged. This data-level design plays a crucial role in allowing Tele-Omni to support both generation and editing behaviors within a single framework.

On the modeling side, Tele-Omni leverages a pretrained multimodal large language model (MLLM) to parse multimodal instructions and infer the required generation or editing behavior. The MLLM does not directly generate videos; instead, it serves as a high-level control module that converts user instructions into structured conditioning signals. Video synthesis is performed by a diffusion-based generator, which receives visual conditions through VAE-based encoders to preserve dense spatial and temporal information. This design avoids the information bottleneck introduced by compressing long video sequences into a small set of semantic tokens, enabling finer-grained control over video content.
By decoupling instruction parsing from video synthesis and combining it with task-aware data processing, Tele-Omni enables flexible multimodal control while maintaining motion continuity and visual consistency across frames. This unified framework allows the model to support diverse video generation and editing scenarios in a consistent and scalable manner.

Experimental results demonstrate that the unified formulation and data design enable Tele-Omni to reliably support multimodal, context-driven video generation and editing within a single system.

\section{Related Works}

\subsection{Multimodal Understanding}
The rapid emergence of multimodal large language models (MLLMs) has substantially advanced multimodal understanding and reasoning~\citep{bai2025qwen2,cao2022multi,li2024llava,guo2025seed1}. To enable complex cross-modal comprehension while preserving strong textual reasoning abilities, early representative studies such as LLaVA and MiniGPT-4 connect modality-specific encoders (e.g., CLIP~\citep{radford2021learning}) with large language models (LLMs)~\citep{achiam2023gpt,dubey2024llama,qwen3} through lightweight projection layers. This design paradigm has inspired a wave of follow-up studies aimed at improving perceptual capacity. Subsequent research explores scaling visual encoders~\citep{chen2024internvl}, supporting higher-resolution inputs~\citep{wang2024qwen2}, adopting efficient Mixture-of-Experts (MoE) architectures~\citep{jiang2024mixtral}, developing more expressive projection modules~\citep{li2023blip,alayrac2022flamingo}, and enhance training recipes~\citep{karamcheti2024prismatic}. For video understanding, a widely adopted approach~\citep{lin2024vila} encodes video frames into sequences of visual tokens that are compatible with LLM inputs. Although these models demonstrate strong multimodal reasoning capabilities, they are fundamentally understanding-oriented and lack the ability to generate multimodal outputs, such as images or videos. 

\subsection{Multimodal Generation}

Early multimodal generation approaches primarily rely on generative adversarial networks (GANs)~\citep{goodfellow2014generative} to synthesize novel visual content. However, GAN-based methods suffer from limited scalability and struggle to produce high-resolution, high-quality results. In contrast, diffusion models~\citep{ho2020denoising} and autoregressive models~\citep{box2015time}, benefiting from stable optimization, and large-scale training based on massive multimodal corpora, have emerged as the dominant paradigms for modern image and video synthesis.

Diffusion models generate data by gradually transforming simple noise distributions into complex data distributions through a sequence of learned denoising steps. Early theoretical work established the foundation of diffusion-based generative modeling, while subsequent studies significantly improved generation quality and efficiency through enhanced training strategies, sampling techniques, and architectural designs~\citep{lu2022dpm,karras2024analyzing,gao2025seedream,seedream2025seedream}. In particular, the introduction of large-scale diffusion frameworks, such as Stable Diffusion and Diffusion Transformers (DiT)~\citep{peebles2023scalable}, has led to substantial progress in high-quality, high-resolution, and long-duration visual generation. Building upon the DiT architecture,  recent works~\citep{zheng2024open,wan2025wan} demonstrate strong performance in text-to-video generation, highlighting the scalability and expressive power of diffusion-based approaches. 

% recent works demonstrate strong performance in instruction-conditioned image and video synthesis, producing visually detailed results that closely align with user inputs.

In parallel, AR models extend the success of sequence modeling in language to the visual domain by tokenizing visual signals into discrete representations and performing next-token prediction over multimodal sequences. In AR models, visual inputs are first encoded into sequences of discrete tokens, after which visual and textual tokens are jointly modeled in an autoregressive manner. To bridge continuous visual signals and discrete token representations, vector quantization techniques~\citep{yu2023language,tian2024visual,li2024autoregressive} are commonly adopted to map high-dimensional data into finite codebooks. Early representative studies~\citep{ramesh2022hierarchical,chen2023class,wang2024causal,betker2023improving} demonstrate that sufficiently scaled AR models can generate high-quality images from textual descriptions. Subsequent studies~\citep{yu2023language} shows that improved visual tokenizers further enhance the performance of AR models in image and video generation. More recent approaches~\citep{tian2024visual,li2024autoregressive,liu2025infinitystar,hu2025omni} extend autoregressive modeling to video synthesis, showing particular strength in modeling complex temporal dynamics and large-scale motions, and explore variants that combine autoregressive prediction with continuous or multi-scale representations.

\subsection{Video Generation and Editing}
Diffusion models have achieved remarkable success in high-fidelity image, video, geometry synthesis and editing~\citep{saharia2022photorealistic,esser2024scaling,podell2023sdxl,zheng2024open,cao20253dot,wan2025wan,yang2025not,shi2023zero123++,cao2025bridging,hu2025auto,poole2022dreamfusion,xiong2025texgaussian,liu2024texoct,liu2025texgarment}. The video generation and editing remain substantially more challenging. In the image domain, controllability has been significantly enhanced through external conditioning mechanisms, such as ControlNet~\citep{zhang2023adding} and T2I-Adapter~\citep{mou2024t2i}. InstructPix2Pix~\citep{brooks2023instructpix2pix} and EMU-Edit~\citep{sheynin2024emu} support instruction-driven editing. Moreover, recent studies~\citep{xiao2025omnigen,tan2025ominicontrol} show a clear trend toward unified image-generation frameworks that extend beyond pure synthesis to support reference-guided and flexible editing. In contrast, the video domain is still largely dominated by task-specific frameworks tailored to either generation or editing, such as Video-P2P~\citep{liu2024video}, MagicEdit~\citep{liew2023magicedit}, and MotionCtrl~\citep{wang2024motionctrl}. Some recent studies, e.g., UniVideo~\citep{wei2025univideo}, integrates MLLMs to guide the video generation and editing process, improving high-level semantic control. Although several efforts have attempted to build more unified video systems, these approaches typically depend on task-dependent designs or specialized components, which limits their flexibility and scalability. As a result, compared with the relatively mature and versatile landscape of image synthesis, achieving a truly unified framework for video generation and editing remains an open and underexplored challenge.
% \subsection{Unified Multimodal Understanding and Generation}
% A concurrent line of research integrates MLLMs to guide the editing process

\section{Methods}
In this section, we describe the design of Tele-Omni, a unified multimodal framework for video generation and video editing. We first introduce the overall model architecture, and then explain how multiple video-centric tasks, including generation and editing are formulated and unified within a single framework through multimodal instruction conditioning.

\begin{figure}[h]
  \centering

  \includegraphics[width=\linewidth]{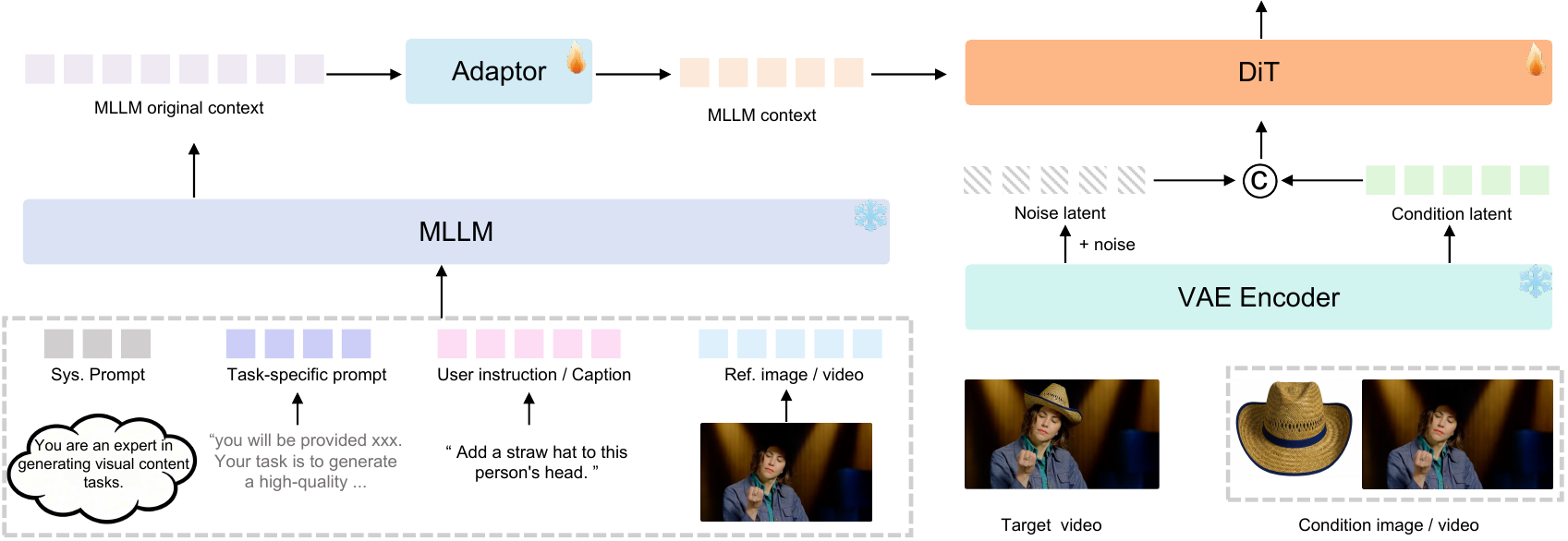}

  \caption{Overview of Tele-Omni. Tele-Omni adopts a two-module architecture, where an MLLM serves as the understanding module and generates editing guidance from the input instruction and visual inputs, and a DiT acts as the generation module connected to the MLLM via an adaptor. During training, the DiT and adaptor are trainable, while the MLLM and VAE are frozen.}
  \label{fig:pipeline}
\end{figure}

\subsection{Overall}

As illustrated in Fig.~\ref{fig:pipeline}, Tele-Omni consists of two core components:
(1) a multimodal large language model (MLLM) for multimodal instruction parsing, and
(2) a diffusion transformer (DiT) that serves as a unified backbone for video generation and video editing.

The MLLM takes text, images, and videos as inputs and is responsible for interpreting user instructions. It encodes multimodal inputs into high-level semantic representations that specify the desired generation or editing behavior. These representations capture key attributes, including appearance constraints, motion intent, and the relationship between reference inputs and the target video.

The DiT receives two types of inputs. First, semantic features extracted from the penultimate hidden state of the MLLM are projected into the conditional space of the DiT via a lightweight, trainable adaptor. These features provide instruction-level guidance and enable flexible, natural language–driven control. Second, task-dependent visual conditions are incorporated, including reference images, reference videos, and first or last frames. All visual conditions are encoded into latent representations using a pretrained VAE~\cite{}.

By jointly conditioning the DiT on semantic and visual signals, Tele-Omni avoids task-specific architectural branches while preserving strong alignment between user intent and generated content. This unified design is particularly important for video editing and frame-based generation, where both high-level semantic consistency and low-level visual fidelity must be maintained.

\subsection{Unified Video Tasks}
Tele-Omni supports multiple video generation and video editing tasks within a single unified framework. Instead of introducing specialized modules for individual tasks, all tasks are formulated through natural language instructions combined with multimodal visual conditions, as illustrated in Fig.~\ref{fig:pipeline}. Task intent is inferred implicitly from the structure and content of the instruction.

\textbf{Text-to-Video Generation.} For text-to-video generation, a textual prompt is provided as the sole instruction. The MLLM parses the text and produces semantic conditioning features that describe the desired content and motion. The DiT takes a noisy video latent as input and progressively denoises it under the guidance of these semantic features, resulting in a video consistent with the textual description. 

% The task-specific prompt is ``\textit{You will receive a video caption as input. Your objective is to produce a high-quality video that precisely represents the described content. Carefully attend to the color, shape, size, texture, quantity, textual details, spatial relationships, and motion of all objects, as well as the characteristics of the background.}".

\textbf{Image-to-Video Generation.} For image-to-video generation, a reference image is provided along with a textual instruction. The MLLM jointly processes the text and image to infer the intended generation behavior, while the reference image is encoded into latent representations by a pretrained VAE and supplied to the DiT as visual conditioning. This formulation allows the generated video to preserve the appearance specified by the reference image, while enabling the model to synthesize motion and temporal dynamics. 

% The task-specific prompt is ``\textit{You will be provided with an input image and a caption. Your task is to generate a high-quality video that preserves visual consistency while faithfully reflecting the content described in the caption.}"

\textbf{In-Context Video Generation.} For in-context generation, multiple visual conditions may be provided, such as a reference image and a reference video. These visual inputs, together with the textual instruction, are first processed by the MLLM to infer the intended generation behavior. Meanwhile, each visual condition is independently encoded into latent representations using the VAE encoder. To accommodate a variable number of visual conditions, all visual latents are aligned to a unified temporal shape and concatenated along the temporal dimension. 

% The task-specific prompt is ``\textit{You will receive a reference image along with a video caption. Your task is to generate a high-quality video that integrates the object from the reference image into a single coherent scene that aligns with the caption.}".

% The DiT processes the concatenated latents using self-attention, allowing it to aggregate information across multiple references and infer how different visual conditions jointly constrain the generated video. This design avoids the need for task-specific bias embeddings or context adapter modules and naturally extends to different numbers and types of visual inputs.

\textbf{In-Context Video Editing.}
For in-context video editing, users provide a reference video together with an editing instruction. The reference video is first encoded by a VAE to obtain latent representations that serve as visual conditions. Meanwhile, both the reference video and the editing instruction are fed into the MLLM, which parses the instruction to identify the target content and its spatial–temporal location for modification. Guided by the parsed instruction and the reference video latents, the DiT denoises the noisy video latents to perform localized or attribute-level edits, while preserving the global motion structure and the unedited regions of the video. 

% The task-specific prompt is ``\textit{You will be given a video and an editing instruction. Your task is to generate a high-quality video by applying the specified edits, ensuring consistency in visual quality, temporal coherence, and alignment with the instruction.}".

\textbf{First–Last-Frame Video Generation}
Tele-Omni also supports video generation conditioned on given start and end frames, which explicitly define the initial and final states of the target video. This setting imposes strong boundary constraints while allowing the model to synthesize intermediate motion flexibly.

Specifically, the start and end frames are treated as standard visual conditions and independently encoded into latent representations using a VAE. These latents are then expanded to a unified temporal shape and concatenated along the temporal dimension. Text instructions can optionally be provided together with the visual conditions to describe the desired transitions or motion characteristics.

During generation, the DiT jointly attends to the semantic guidance from the MLLM and the visual constraints imposed by the start and end frame latents. As a result, Tele-Omni generates temporally coherent videos with smooth transitions between the specified boundary frames, without introducing task-specific mechanisms or architectural modifications.  

% The task-specific prompt is ``\textit{You will be given a starting frame and an ending frame. Your task is to generate a high-quality video that creates a seamless and logically coherent transition from the starting frame to the ending frame. Ensure temporal continuity and preserve the color, shape, texture, quantity, text, and spatial relationships of all objects and the background throughout the transition, while naturally and logically depicting the evolution between the two frames.}".

\subsection{Conditioning and Positional Encoding}
IC-Lora~\citep{huang2024context} and Omni-Transfer~\citep{zhang2026omnitransfer} demonstrate that in-context generation inherently exhibits superior performance in text-to-video and video-to-video generation tasks. Building on this, we generalize this strategy to our unified video generation tasks. For the target latent, the position embedding follows the standard 3D RoPE, defined as $R_\theta(\Delta=(0,0,0))$, where $\Delta$ denotes the applied offset. In the case of video-conditioned video generation, the RoPE for the condition latent is spatially shifted as $R_\theta(\Delta=(0,w_{tar},0))$, where $w_{tar}$ refers to width of target latents. Conversely, for first-last-frame conditioning and image-to-video tasks, the embedding reduces to the standard 3D RoPE.Specifically for the in-context generation task, the RoPE embedding is formulated as:
\[R_\theta =
\begin{cases}
R_\theta\!\left(\Delta = (f+1,0, 0)\right),
& \text{for reference image}, \\[6pt]
R_\theta\!\left(\Delta = (0, w_{tar}, 0)\right),
& \text{for conditional video}.
\end{cases}
\]
Here, both reference images and conditional videos are concatenated along the token dimension. The conditional videos function as spatial in-context conditions (shifted by width), while reference images serve as temporal conditions (shifted by time).
% To enable the DiT to correctly interpret multiple visual conditions and distinguish them from the noisy video latent to be generated, we apply 3D positional embeddings to all visual tokens. Spatial indices are preserved across frames, while the temporal index is incremented for each concatenated visual input.

% This positional encoding scheme provides a consistent temporal ordering for heterogeneous visual conditions, including reference images, reference videos, and boundary frames. In practice, it allows the DiT to effectively leverage contextual and boundary information, and we find it more reliable than approaches that offset all spatial and temporal axes whenever a new visual input is introduced.

\subsection{Training strategies}
The training process is divided into two stages:
Stage 1. Adapter alignment.
In the first stage, only the adapter is optimized, while both the MLLM and the DiT remain frozen. Training is performed on image-to-video and text-to-video generation tasks, enabling the adapter to effectively bridge the semantic representations produced by the MLLM with the video generation process. Upon completion of this stage, TeleOmni can generate videos conditioned on semantic feature of the MLLM.
Stage 2. Joint adapter–DiT fine-tuning.
In the second stage, the training data are expanded to include in-context generation, in-context video editing, and first–last-frame video generation tasks, in addition to the original image-to-video and text-to-video generation tasks. During this stage, the MLLM remains frozen, while the adapter and the DiT are jointly optimized, leading to improved temporal coherence, stronger instruction adherence, and enhanced editing fidelity.

\section{Data System}

\subsection{Instruction based video editing}

We categorize instruction-based video editing tasks into two broad classes: global editing and local editing. Global editing tasks modify all pixels in the video, encompassing tasks such as \textit{style transfer}. In contrast, local editing tasks are restricted to specific spatial regions and primarily involve \textit{inserting}, \textit{removing}, or \textit{modifying} a subject in the video.

\noindent \textbf{Style transfer.} This task focuses on modifying the visual style of a video while preserving its original motion dynamics and low-level texture details. In the data generation pipeline, the first frame of the video is supplied to GPT-4o together with a textual prompt, which is used to generate corresponding instructions for both image and video editing. The first frame and the resulting image-editing instruction are then passed to FLUX-Kontext-dev~\citep{labs2025flux1kontextflowmatching} to obtain a stylistically transformed version of the first frame. Subsequently, this edited first frame, along with Canny edge maps or a depth sequence extracted from the original video using OpenCV or Video Depth Anything~\citep{video_depth_anything}, is provided as input to the Wan2.1-Fun-V1.1-14B-Control~\citep{wan2025wan}, which synthesizes a new video that preserves the semantic content and motion structure of the source while exhibiting a distinct visual style. We collect 18 commonly used artistic styles, including artistic styles (e.g., Oil painting, Post-Impressionism), anime styles (e.g., Chibi), and weather conditions (e.g., Sunny, Rainy, Snowy).

\noindent \textbf{Insertion.} For the subject-insertion task, we first employ DiffuEraser~\citep{li2025diffueraserdiffusionmodelvideo} to remove the target subject from the original video clip given a predefined mask. The resulting inpainted video, in which the subject is absent, is then used as the source video for the insertion task. Subsequently, we utilize GPT-4o to generate a corresponding subject-insertion instruction conditioned on the textual description of the removed subject, while the original, unedited video serves as the ground-truth.

\noindent \textbf{Removal.} Conversely, to generate paired data for the subject-removal task, we synthesize a video containing the target subject and treat this synthesized video as the source video. Specifically, we first prompt GPT-4o to generate a textual description of the subject to be inserted into the scene, which is appended to the first frame of the original video as conditioning metadata. This first frame is then edited using the FLUX-Kontext-dev~\citep{labs2025flux1kontextflowmatching} to insert the described subject. Next, Wan2.2-I2V-A14B~\citep{wan2025wan} is used to generate a full video clip conditioned on the edited first frame. We then segment the newly inserted subject in the generated video using the Grounded SAM2~\citep{ren2024grounded,ravi2024sam2}. The resulting subject mask is used to extract the corresponding region, which is composited onto the original video frames, yielding the final edited video that contains the added subject. This synthesized video subsequently serves as the source video for the subject-removal task. The subject descriptions produced by GPT-4o are also reused to formulate the textual prompts for this task.

\noindent \textbf{Modifying.} We consider two types of modification settings: coarse-grained, such as the background modification, and fine-grained, such as the subject modification. In the former setting, we first filter videos that exhibit a clearly separable foreground and background, following OpenVE~\citep{openve}. We then employ GPT-4o to automatically generate textual prompts that specify the desired background modification. Given these prompts, FLUX-Kontext-dev~\citep{labs2025flux1kontextflowmatching} is used to replace the background in the first video frame. Subsequently, the Canny and/or depth maps of the original video, together with the edited first frame, are provided as input to the Wan2.1-Fun-V1.1-14B-Control~\citep{wan2025wan} to synthesize the final video with the modified background. Throughout this pipeline, we additionally provide a foreground mask corresponding to the primary subject as input to both the first-frame editing stage and the subsequent video generation stage, thereby ensuring that the foreground in the synthesized video remains consistent with that in the original sequence. In the latter setting, the procedure is analogous to the subject insertion and removal pipeline. Specifically, we input the initial video frame and the textual description of the target subject into GPT-4o to obtain video-editing prompts. FLUX-Kontext-dev~\citep{labs2025flux1kontextflowmatching} then performs localized editing on the first frame. Finally, the edited frame, together with the original Canny or depth video, is passed to Wan2.1-Fun-V1.1-14B-Control~\citep{wan2025wan} to generate the edited video clip.

\begin{figure}[!h]
  \centering
  \includegraphics[width=\linewidth]{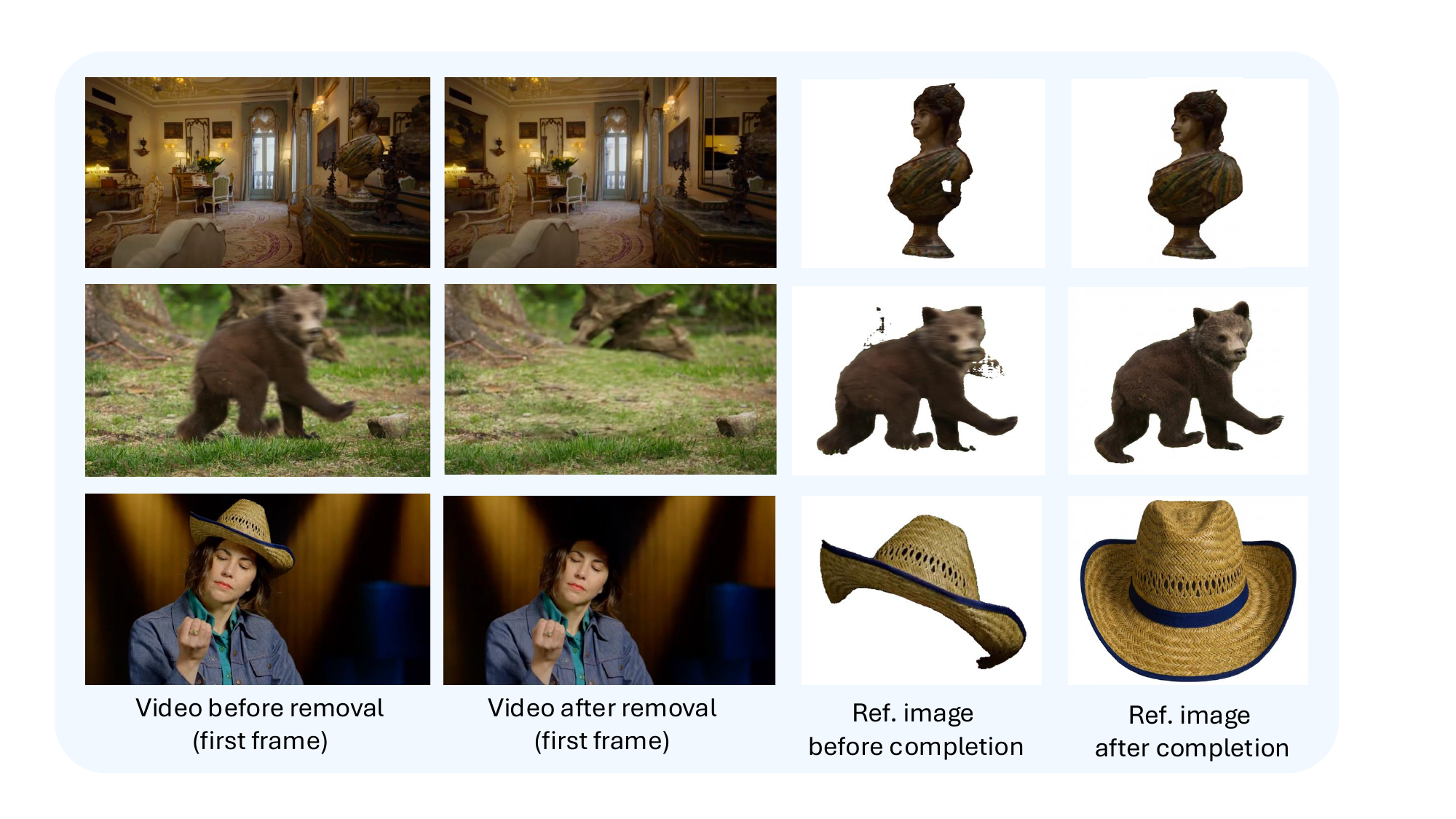}
  \caption{Samples of constructed image-guided video editing dataset}
  \label{fig:i+v2v_data}
\end{figure}

\subsection{Image based object insertion and removal data generation}
We design a robust pipeline to generate high-quality training pairs for image-guided video object insertion and removal task. Utilizing date generated by above section, our process involves three key stages:

\begin{itemize}
    \item Object Selection: We first employ Gemini3-pro~\citep{gemini} to automatically identify differences between input video and GT video together with editing prompt to identify the edited objects.
    \item Video Processing: After obtaining objects descriptions, we further use LangSAM~\citep{langsam} to segment out the edited objects. For previously constructed dataset, we use input video for insertion task and GT video for removal task to do the segmentations.
    \item Reference Generation Refinement: Due to occlusions in videos, the raw object segments extracted by LangSAM are often incomplete. To address this, we process the segmented raw images using Nano-Banana to reconstruct the missing regions. This results in high-quality, complete reference images displayed against a white background.
\end{itemize}

Finally, we mix all the data generated from insertion and removal dataset and construct the image guided raw video dataset. 
To ensure high quality, we filter out samples that exhibit any of the following artifacts:
(1) Identity inconsistency: Significant identity changes in the reference image after completion.
(2) Incomplete removal: Residual object parts or ghosting artifacts in the video.
(3) Unnatural inpainting: Visual anomalies in the inpainted video regions.
(4) Unintended editing: Unintended modifications to regions outside the target object.
We employ a dual-verification strategy using Qwen3-VL-32B~\citep{qwen3} and Gemini3-pro~\citep{gemini}. Only samples that pass the strict assessment of both models are retained. We show data samples in figure~\ref{fig:i+v2v_data}

% \begin{figure}
%   \centering
%   % 2. 开始画有背景的盒子
%   % colback=背景色, colframe=边框色(设为背景色即无边框), arc=圆角大小
%   \begin{tcolorbox}[
%     colback=techblue, 
%     colframe=techblue, 
%     arc=10pt, 
%     boxsep=5pt,      % 内容和边框的距离
%     width=\linewidth % 盒子占满行宽
%   ]
%   \includegraphics[scale=0.6]{imgs/i+v2v_data.pdf}
%     \begin{tabular}{ 
%         c
%         >{\centering\arraybackslash}p{2.7cm}  % 第1列
%         c      
%         >{\centering\arraybackslash}p{2.7cm}  % 第2列
%         c                                     % 第3列（中间的大空隙）
%         >{\centering\arraybackslash}p{2.8cm}  % 第4列
%         c      
%         >{\centering\arraybackslash}p{2.8cm}  % 第5列
%       }
%         % --- 表格内容 ---
        
%         % 第一列内容
           
%         \hspace{0.5cm} 
%         &
%         Video before process 
%         & 
%         \hspace{1.2cm} 
%         &
%         % 第二列内容
%         Video after process 
%         & 
%         % 第三列（空隙），用 \hspace 撑开宽度
%         \hspace{0.5cm} 
%         & 
%         % 第四列内容
%         Ref. image before completion 
%         &
%         \hspace{0.2cm} 
%         & 
%         % 第五列内容
%         Ref. image after completion 
%       \end{tabular}

%   \end{tcolorbox} % 盒子结束
%   \caption{Samples of constructed dataset}
%   \label{fig:i+v2v_data}
% \end{figure}

\section{Experiments and Discussion}
\begin{figure}[!th]
  \centering
  \includegraphics[width=\linewidth]{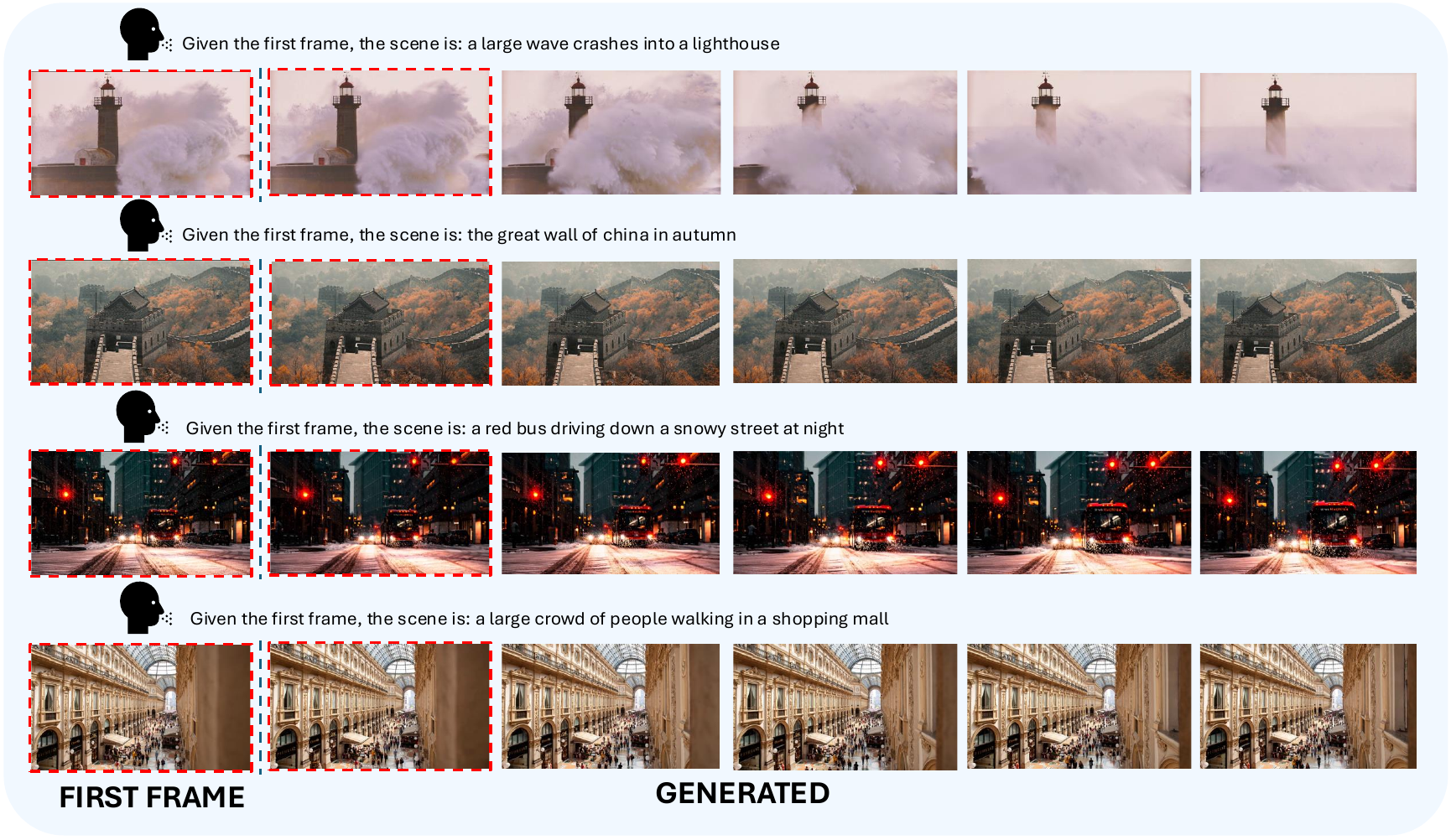}
  \caption{Example generated videos for image and text prompted video generation task.}
  \label{fig:i2v_result}
\end{figure}

\subsection{Image to Video Generation}
Figure~\ref{fig:i2v_result} presents the image-to-video results of our Tele-Omni. Given a single image prompt, Tele-Omni generates videos that are visually coherent and aesthetically pleasing. Evaluations on diverse video examples show that the generated results exhibit pronounced action dynamics, effectively avoiding overly static or frozen motions. As illustrated by the first, third, and last videos in Figure~\ref{fig:i2v_result}, our method produces strong motion dynamics across different motion types.

In particular, the ocean waves in the first video and the falling snow in the third video display vivid and continuous \textbf{particle dynamics}, while the crowd movement in the last video highlights the model’s ability to generate robust and general \textbf{motion dynamics} beyond particle-based motions. Moreover, the generated videos are free from noticeable blurriness and visual artifacts.

\subsection{First and Last Frame Conditioned Video Generation}
\begin{figure}[!h]
  \centering
  \includegraphics[width=\linewidth]{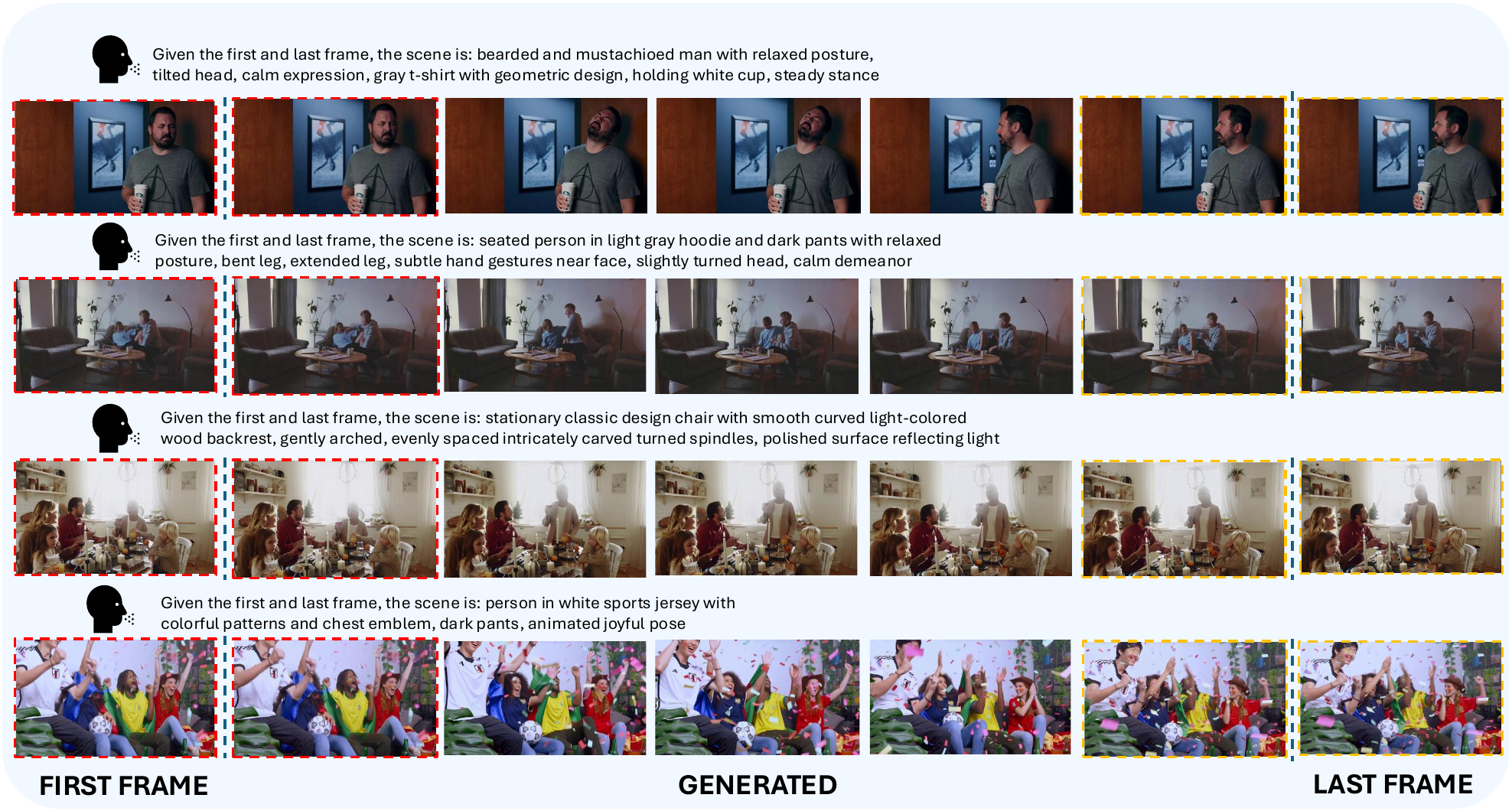}
  \caption{Example generated videos for first and last frame conditioned video generation task.}
  \label{fig:fl2v_result}
\end{figure}

Figure~\ref{fig:fl2v_result} presents the first-last-frame-to-video results of our Tele-Omni. The generated videos faithfully follow the content of both the first and the last conditioning frames. The intermediate frames exhibit smooth and coherent transitions, avoiding abrupt motion changes or sudden jumps from the initial condition to the final frame. As a result, the overall temporal evolution is natural and visually plausible.

More importantly, as shown in the last example, under first- and last-frame conditioning, Tele-Omni is able to correctly interpret the falling ribbon motion in the video and plausibly complete the intermediate frames. This indicates that the model can reason about complex motion evolution rather than simply interpolating appearances. In addition, Tele-Omni demonstrates a correct understanding of lighting and occlusion relationships. For example, in the third case, when the woman stands up, the resulting changes in light occlusion caused by the human body are accurately synthesized.

\begin{figure}[!th]
  \centering
  \includegraphics[width=\linewidth]{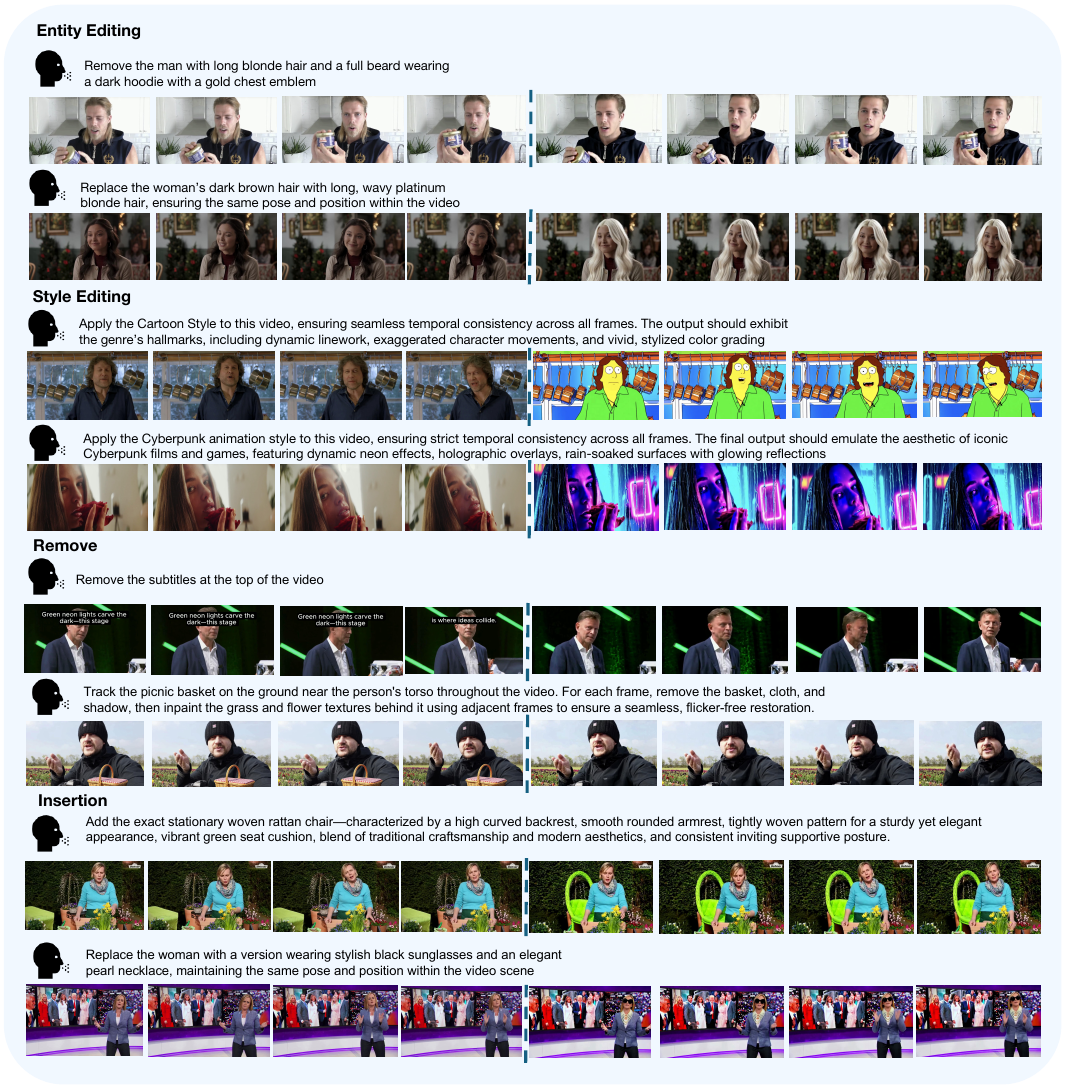}
  \caption{Example generated videos for video editing task. This task is further divided into four categories: (1) Entity Editing, (2) Style Editing, (3) Remove and (4) Insertion.}
  \label{fig:v2v_result}
\end{figure}
\subsection{Video Editing}
Figure~\ref{fig:v2v_result} presents the video editing results of our Tele-Omni. 
In this section, we categorize video editing tasks into four sub-tasks: entity editing, style editing, object removal, and object insertion.
For entity editing, Tele-Omni focuses on modifying foreground entities while preserving their identity and maintaining consistency with the surrounding scene. The edited entities remain coherent in both appearance and motion throughout the video, demonstrating robust temporal consistency.
For style editing, our model does not rigidly transform the appearance based on the original object geometry. Instead, Tele-Omni adapts the overall visual style by jointly reasoning over the textual instruction and the video content, enabling substantial stylistic changes while preserving the identity of both the entities and the scene.
For object removal, Tele-Omni produces clean results without introducing ghosting artifacts. The removed regions are plausibly inpainted with high-quality details that are consistent with the surrounding context, resulting in visually seamless outputs.
For object insertion, our examples extend beyond static objects. Dynamically inserted entities interact naturally with the scene over time, demonstrating the model’s ability to model scene dynamics and maintain coherent entity–scene interactions.
\begin{figure}[!th]
  \centering
  \includegraphics[width=\linewidth,height=7cm]{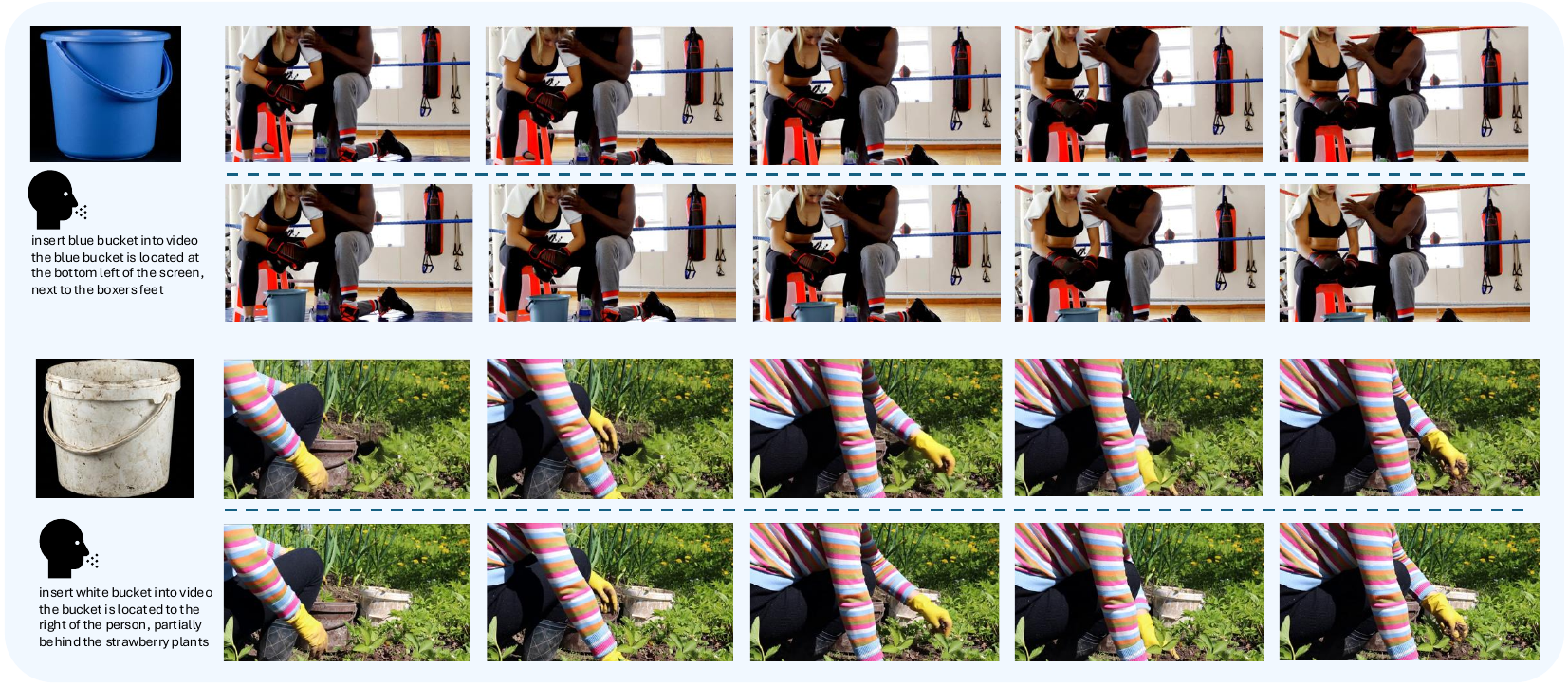}
  \caption{Example generated videos for In-context editing task.}
  \label{fig:i+v2v_result}
\end{figure}

\begin{figure}[!h]
  \centering
  \includegraphics[width=\linewidth,height=3.3cm]{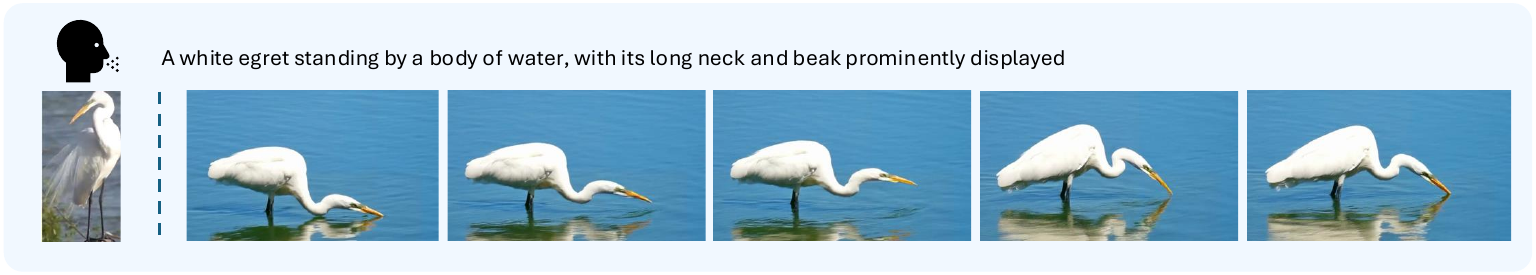}
  \caption{Example generated videos for In-context generation task.}
  \label{fig:in_context_gen}
\end{figure}

\begin{figure}[!h]
  \centering
  \includegraphics[width=\linewidth,height=13cm]{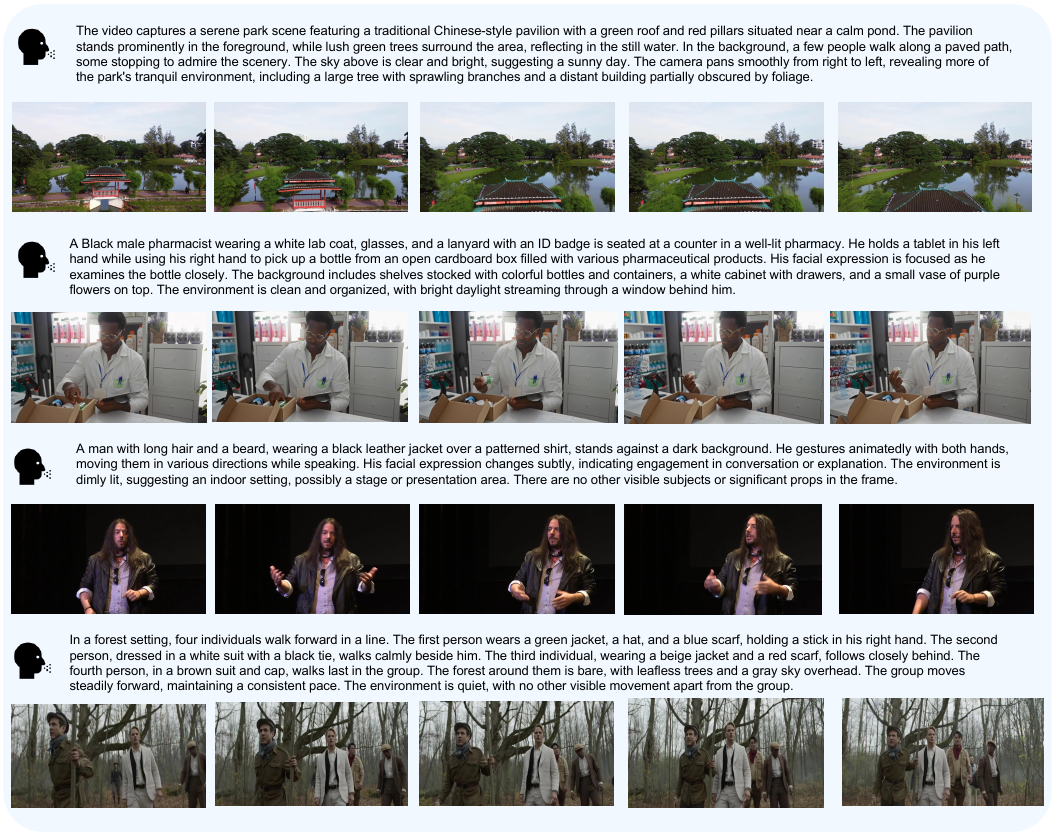}
  \caption{Example generated videos for text to video generation task.}
  \label{fig:t2v_result}
\end{figure}
\subsection{In-Context Editing }
Figure~\ref{fig:i+v2v_result} presents the in-context editing results of our Tele-Omni. Given a conditioning image, a target video, and a textual instruction, Tele-Omni is able to insert the image content into the video in a mask-free manner at the specified location. The inserted object remains spatially stable over time, preserving its geometry and appearance without noticeable deformation. Moreover, the object does not drift from the designated position across frames, demonstrating strong temporal consistency in both placement and structure.

\subsection{In-Context Generation }
As shown in Figure~\ref{fig:in_context_gen}, Tele-Omni excels in in-context generation. Conditioned on a reference image and a text prompt, the model successfully maintains high identity fidelity for the subject. Crucially, the generated videos modify the background according to the textual instruction rather than overfitting to the reference image. This confirms that our MLLM embeddings effectively encode task-specific prompts to control the generation context.

\subsection{Text to Video Generation}
Figure~\ref{fig:t2v_result} presents the text-to-video results of our Tele-Omni. Text-to-video generation is a fundamental capability in video generation. Our results show that Tele-Omni retains the text-to-video generation quality of existing foundation models while extending them with unified video generation and editing capabilities. All generated videos display coherent motion dynamics without introducing noticeable artifacts.

\section{Conclusion}
We introduced Tele-Omni, a versatile model designed to unify text-to-video, image-to-video, and complex editing tasks such as in-context editing and frame-conditioned generation. Our architecture effectively combines the semantic understanding of advanced VLMs with a video generator through adapter-based fine-tuning. The resulting model exhibits exceptional scene understanding and generation quality, producing consistent and artifact-free videos. Tele-Omni stands as a promising foundation model, paving the way for next-generation video synthesis applications.

\section*{Contributors}

\textbf{Project Leaders:} Haibin Huang, Chi Zhang, Xuelong Li

\noindent \textbf{Core Contributors:} Jialun Liu, Tian Li, Xiao Cao, Yukuo Ma, Gonghu Shang

\noindent \textbf{Contributors (alphabet order):} 
Xiangzhen Chang, Zhiyong Huang, Jiakui Hu, Zuoxin Li, Yuanzhi Liang, Cong Liu, Junqi Liu, Robby T. Tan, Haitong Tang, Qizhen Weng, Yifan Xu, Liying Yang, Xiaoyan Yang, Peng Yu, Shiwen Zhang
% \noindent \textbf{Contributors (alphabet order):} Cong Liu, Junqi Liu, Jiakui Hu, Robby T. Tan, Shiwen Zhang, Liying Yang, Xiaoyan Yang, Qizhen Weng, Xiangzhen Chang, Yuanzhi Liang, Yifan Xu, Zhiyong Huang, Zuoxin Li, Peng Yu, Haitong Tang
% \textbf{Project Leaders}: Haibin Huang, Chi Zhang, Xuelong Li \\ \textbf{Core Contributors}: Jialun Liu, Yukuo Ma, Xiao Cao, Tian Li, Gonghu Shang \\ \textbf{Contributors} (listed alphabetically): 
% Cong Liu,
% Junqi Liu,
% Jiakui Hu,
% Robby T. Tan,
% Shiwen Zhang,
% Liying Yang,
% Xiaoyan Yang,
% Qizhen Weng,
% Xiangzhen Chang,
% Yuanzhi Liang,
% Yifan Xu,
% Zhiyong Huang,
% Zuoxin Li
% }

% \authorlist{

% \paragraph{\textbf{Contributors (Listed alphabetically):}} Cong Liu, Junqi Liu, Jiakui Hu, Robby T. Tan, Shiwen Zhang, Liying Yang, Xiaoyan Yang, Qizhen Weng, Xiangzhen Chang, Yuanzhi Liang, Yifan Xu, Zhiyong Huang, Zuoxin Li 

% }

\clearpage

\bibliography{paper}

\begin{thebibliography}{}

\bibitem[\protect\citename{lab, }n.d.]{labs2025flux1kontextflowmatching}
{\em FLUX.1 Kontext: Flow Matching for In-Context Image Generation and Editing in Latent Space}.

\bibitem[\protect\citename{Achiam {\em et~al.}, }2023]{achiam2023gpt}
Achiam, Josh, Adler, Steven, Agarwal, Sandhini, Ahmad, Lama, Akkaya, Ilge, Aleman, Florencia~Leoni, Almeida, Diogo, Altenschmidt, Janko, Altman, Sam, Anadkat, Shyamal, {\em et~al.} 2023.
\newblock Gpt-4 technical report.
\newblock {\em arXiv preprint arXiv:2303.08774}.

\bibitem[\protect\citename{Alayrac {\em et~al.}, }2022]{alayrac2022flamingo}
Alayrac, Jean-Baptiste, Donahue, Jeff, Luc, Pauline, Miech, Antoine, Barr, Iain, Hasson, Yana, Lenc, Karel, Mensch, Arthur, Millican, Katherine, Reynolds, Malcolm, {\em et~al.} 2022.
\newblock Flamingo: a visual language model for few-shot learning.
\newblock {\em Advances in neural information processing systems}, {\bf 35}, 23716--23736.

\bibitem[\protect\citename{Bai {\em et~al.}, }2023]{bai2023qwen}
Bai, Jinze, Bai, Shuai, Chu, Yunfei, Cui, Zeyu, Dang, Kai, Deng, Xiaodong, Fan, Yang, Ge, Wenbin, Han, Yu, Huang, Fei, {\em et~al.} 2023.
\newblock Qwen technical report.
\newblock {\em arXiv preprint arXiv:2309.16609}.

\bibitem[\protect\citename{Bai {\em et~al.}, }2025a]{bai2025qwen2}
Bai, Shuai, Chen, Keqin, Liu, Xuejing, Wang, Jialin, Ge, Wenbin, Song, Sibo, Dang, Kai, Wang, Peng, Wang, Shijie, Tang, Jun, {\em et~al.} 2025a.
\newblock Qwen2. 5-vl technical report.
\newblock {\em arXiv preprint arXiv:2502.13923}.

\bibitem[\protect\citename{Bai {\em et~al.}, }2025b]{qwen3}
Bai, Shuai, Cai, Yuxuan, Chen, Ruizhe, Chen, Keqin, Chen, Xionghui, Cheng, Zesen, Deng, Lianghao, Ding, Wei, Gao, Chang, Ge, Chunjiang, Ge, Wenbin, Guo, Zhifang, Huang, Qidong, Huang, Jie, Huang, Fei, Hui, Binyuan, Jiang, Shutong, Li, Zhaohai, Li, Mingsheng, Li, Mei, Li, Kaixin, Lin, Zicheng, Lin, Junyang, Liu, Xuejing, Liu, Jiawei, Liu, Chenglong, Liu, Yang, Liu, Dayiheng, Liu, Shixuan, Lu, Dunjie, Luo, Ruilin, Lv, Chenxu, Men, Rui, Meng, Lingchen, Ren, Xuancheng, Ren, Xingzhang, Song, Sibo, Sun, Yuchong, Tang, Jun, Tu, Jianhong, Wan, Jianqiang, Wang, Peng, Wang, Pengfei, Wang, Qiuyue, Wang, Yuxuan, Xie, Tianbao, Xu, Yiheng, Xu, Haiyang, Xu, Jin, Yang, Zhibo, Yang, Mingkun, Yang, Jianxin, Yang, An, Yu, Bowen, Zhang, Fei, Zhang, Hang, Zhang, Xi, Zheng, Bo, Zhong, Humen, Zhou, Jingren, Zhou, Fan, Zhou, Jing, Zhu, Yuanzhi, \& Zhu, Ke. 2025b.
\newblock {\em Qwen3-VL Technical Report}.

\bibitem[\protect\citename{Betker {\em et~al.}, }2023]{betker2023improving}
Betker, James, Goh, Gabriel, Jing, Li, Brooks, Tim, Wang, Jianfeng, Li, Linjie, Ouyang, Long, Zhuang, Juntang, Lee, Joyce, Guo, Yufei, {\em et~al.} 2023.
\newblock Improving image generation with better captions.
\newblock {\em Computer Science. https://cdn. openai. com/papers/dall-e-3. pdf}, {\bf 2}(3), 8.

\bibitem[\protect\citename{Box {\em et~al.}, }2015]{box2015time}
Box, George~EP, Jenkins, Gwilym~M, Reinsel, Gregory~C, \& Ljung, Greta~M. 2015.
\newblock {\em Time series analysis: forecasting and control}.
\newblock John Wiley \& Sons.

\bibitem[\protect\citename{Brooks {\em et~al.}, }2023]{brooks2023instructpix2pix}
Brooks, Tim, Holynski, Aleksander, \& Efros, Alexei~A. 2023.
\newblock Instructpix2pix: Learning to follow image editing instructions.
\newblock {\em Pages  18392--18402 of:} {\em Proceedings of the IEEE/CVF conference on computer vision and pattern recognition}.

\bibitem[\protect\citename{Cao {\em et~al.}, }n.d.]{cao2025bridging}
Cao, Xiao, Zhao, Yuyang, Tan, Robby~T, \& Huang, Zhiyong.
\newblock Bridging 3D Editing and Geometry-Consistent Paired Dataset Creation for 2D Nighttime-to-Daytime Translation.
\newblock {\em In:} {\em [CVPR 2025 Workshop] SyntaGen: 2nd Workshop on Harnessing Generative Models for Synthetic Visual Datasets}.

\bibitem[\protect\citename{Cao {\em et~al.}, }2022]{cao2022multi}
Cao, Xiao, Chen, Zitan, Le, Canyu, \& Meng, Lei. 2022.
\newblock Multi-modal video chapter generation.
\newblock {\em arXiv preprint arXiv:2209.12694}.

\bibitem[\protect\citename{Cao {\em et~al.}, }2025]{cao20253dot}
Cao, Xiao, Lin, Beibei, Wang, Bo, Huang, Zhiyong, \& Tan, Robby~T. 2025.
\newblock 3DOT: Texture Transfer for 3DGS Objects from a Single Reference Image.
\newblock {\em In:} {\em The Thirty-ninth Annual Conference on Neural Information Processing Systems}.

\bibitem[\protect\citename{Chen {\em et~al.}, }2025]{video_depth_anything}
Chen, Sili, Guo, Hengkai, Zhu, Shengnan, Zhang, Feihu, Huang, Zilong, Feng, Jiashi, \& Kang, Bingyi. 2025.
\newblock Video Depth Anything: Consistent Depth Estimation for Super-Long Videos.
\newblock {\em arXiv:2501.12375}.

\bibitem[\protect\citename{Chen {\em et~al.}, }2024]{chen2024internvl}
Chen, Zhe, Wu, Jiannan, Wang, Wenhai, Su, Weijie, Chen, Guo, Xing, Sen, Zhong, Muyan, Zhang, Qinglong, Zhu, Xizhou, Lu, Lewei, {\em et~al.} 2024.
\newblock Internvl: Scaling up vision foundation models and aligning for generic visual-linguistic tasks.
\newblock {\em Pages  24185--24198 of:} {\em Proceedings of the IEEE/CVF conference on computer vision and pattern recognition}.

\bibitem[\protect\citename{Chen {\em et~al.}, }2023]{chen2023class}
Chen, Zitan, Qi, Zhuang, Cao, Xiao, Li, Xiangxian, Meng, Xiangxu, \& Meng, Lei. 2023.
\newblock Class-level structural relation modeling and smoothing for visual representation learning.
\newblock {\em Pages  2964--2972 of:} {\em Proceedings of the 31st ACM International Conference on Multimedia}.

\bibitem[\protect\citename{Dubey {\em et~al.}, }2024]{dubey2024llama}
Dubey, Abhimanyu, Jauhri, Abhinav, Pandey, Abhinav, Kadian, Abhishek, Al-Dahle, Ahmad, Letman, Aiesha, Mathur, Akhil, Schelten, Alan, Yang, Amy, Fan, Angela, {\em et~al.} 2024.
\newblock The llama 3 herd of models.
\newblock {\em arXiv e-prints},  arXiv--2407.

\bibitem[\protect\citename{Esser {\em et~al.}, }2024]{esser2024scaling}
Esser, Patrick, Kulal, Sumith, Blattmann, Andreas, Entezari, Rahim, M{\"u}ller, Jonas, Saini, Harry, Levi, Yam, Lorenz, Dominik, Sauer, Axel, Boesel, Frederic, {\em et~al.} 2024.
\newblock Scaling rectified flow transformers for high-resolution image synthesis.
\newblock {\em In:} {\em Forty-first international conference on machine learning}.

\bibitem[\protect\citename{Gao {\em et~al.}, }2025]{gao2025seedream}
Gao, Yu, Gong, Lixue, Guo, Qiushan, Hou, Xiaoxia, Lai, Zhichao, Li, Fanshi, Li, Liang, Lian, Xiaochen, Liao, Chao, Liu, Liyang, {\em et~al.} 2025.
\newblock Seedream 3.0 technical report.
\newblock {\em arXiv preprint arXiv:2504.11346}.

\bibitem[\protect\citename{Goodfellow {\em et~al.}, }2014]{goodfellow2014generative}
Goodfellow, Ian~J, Pouget-Abadie, Jean, Mirza, Mehdi, Xu, Bing, Warde-Farley, David, Ozair, Sherjil, Courville, Aaron, \& Bengio, Yoshua. 2014.
\newblock Generative adversarial nets.
\newblock {\em Advances in neural information processing systems}, {\bf 27}.

\bibitem[\protect\citename{Guo {\em et~al.}, }2025]{guo2025seed1}
Guo, Dong, Wu, Faming, Zhu, Feida, Leng, Fuxing, Shi, Guang, Chen, Haobin, Fan, Haoqi, Wang, Jian, Jiang, Jianyu, Wang, Jiawei, {\em et~al.} 2025.
\newblock Seed1. 5-vl technical report.
\newblock {\em arXiv preprint arXiv:2505.07062}.

\bibitem[\protect\citename{Gupta {\em et~al.}, }2024]{gupta2024photorealistic}
Gupta, Agrim, Yu, Lijun, Sohn, Kihyuk, Gu, Xiuye, Hahn, Meera, Li, Fei-Fei, Essa, Irfan, Jiang, Lu, \& Lezama, Jos{\'e}. 2024.
\newblock Photorealistic video generation with diffusion models.
\newblock {\em Pages  393--411 of:} {\em European Conference on Computer Vision}.
\newblock Springer.

\bibitem[\protect\citename{He {\em et~al.}, }2025]{openve}
He, Haoyang, Wang, Jie, Zhang, Jiangning, Xue, Zhucun, Bu, Xingyuan, Yang, Qiangpeng, Wen, Shilei, \& Xie, Lei. 2025.
\newblock OpenVE-3M: A Large-Scale High-Quality Dataset for Instruction-Guided Video Editing.
\newblock {\em arXiv preprint arXiv:2512.07826}.

\bibitem[\protect\citename{Ho {\em et~al.}, }2020]{ho2020denoising}
Ho, Jonathan, Jain, Ajay, \& Abbeel, Pieter. 2020.
\newblock Denoising diffusion probabilistic models.
\newblock {\em Advances in neural information processing systems}, {\bf 33}, 6840--6851.

\bibitem[\protect\citename{Hong {\em et~al.}, }2023]{Cogvideo}
Hong, Wenyi, Ding, Ming, Zheng, Wendi, Liu, Xinghan, \& Tang, Jie. 2023.
\newblock CogVideo: Large-scale Pretraining for Text-to-Video Generation via Transformers.
\newblock {\em In:} {\em ICLR}.

\bibitem[\protect\citename{Hu {\em et~al.}, }2025a]{hu2025auto}
Hu, JiaKui, Yang, Yuxiao, Liu, Jialun, Wu, Jinbo, Zhao, Chen, \& Lu, Yanye. 2025a.
\newblock Auto-Regressively Generating Multi-View Consistent Images.
\newblock {\em arXiv preprint arXiv:2506.18527}.

\bibitem[\protect\citename{Hu {\em et~al.}, }2025b]{hu2025omni}
Hu, JiaKui, Zhao, Shanshan, Chen, Qing-Guo, Qiu, Xuerui, Liu, Jialun, Xu, Zhao, Luo, Weihua, Zhang, Kaifu, \& Lu, Yanye. 2025b.
\newblock Omni-View: Unlocking How Generation Facilitates Understanding in Unified 3D Model based on Multiview images.
\newblock {\em arXiv preprint arXiv:2511.07222}.

\bibitem[\protect\citename{Huang {\em et~al.}, }2024]{huang2024context}
Huang, Lianghua, Wang, Wei, Wu, Zhi-Fan, Shi, Yupeng, Dou, Huanzhang, Liang, Chen, Feng, Yutong, Liu, Yu, \& Zhou, Jingren. 2024.
\newblock In-context lora for diffusion transformers.
\newblock {\em arXiv preprint arXiv:2410.23775}.

\bibitem[\protect\citename{Jiang {\em et~al.}, }2024]{jiang2024mixtral}
Jiang, Albert~Q, Sablayrolles, Alexandre, Roux, Antoine, Mensch, Arthur, Savary, Blanche, Bamford, Chris, Chaplot, Devendra~Singh, Casas, Diego de~las, Hanna, Emma~Bou, Bressand, Florian, {\em et~al.} 2024.
\newblock Mixtral of experts.
\newblock {\em arXiv preprint arXiv:2401.04088}.

\bibitem[\protect\citename{Jiang {\em et~al.}, }2025]{jiang2025vace}
Jiang, Zeyinzi, Han, Zhen, Mao, Chaojie, Zhang, Jingfeng, Pan, Yulin, \& Liu, Yu. 2025.
\newblock Vace: All-in-one video creation and editing.
\newblock {\em arXiv preprint arXiv:2503.07598}.

\bibitem[\protect\citename{Karamcheti {\em et~al.}, }2024]{karamcheti2024prismatic}
Karamcheti, Siddharth, Nair, Suraj, Balakrishna, Ashwin, Liang, Percy, Kollar, Thomas, \& Sadigh, Dorsa. 2024.
\newblock Prismatic vlms: Investigating the design space of visually-conditioned language models.
\newblock {\em In:} {\em Forty-first International Conference on Machine Learning}.

\bibitem[\protect\citename{Karras {\em et~al.}, }2024]{karras2024analyzing}
Karras, Tero, Aittala, Miika, Lehtinen, Jaakko, Hellsten, Janne, Aila, Timo, \& Laine, Samuli. 2024.
\newblock Analyzing and improving the training dynamics of diffusion models.
\newblock {\em Pages  24174--24184 of:} {\em Proceedings of the IEEE/CVF Conference on Computer Vision and Pattern Recognition}.

\bibitem[\protect\citename{Li {\em et~al.}, }2024a]{li2024llava}
Li, Bo, Zhang, Yuanhan, Guo, Dong, Zhang, Renrui, Li, Feng, Zhang, Hao, Zhang, Kaichen, Zhang, Peiyuan, Li, Yanwei, Liu, Ziwei, {\em et~al.} 2024a.
\newblock Llava-onevision: Easy visual task transfer.
\newblock {\em arXiv preprint arXiv:2408.03326}.

\bibitem[\protect\citename{Li {\em et~al.}, }2023]{li2023blip}
Li, Junnan, Li, Dongxu, Savarese, Silvio, \& Hoi, Steven. 2023.
\newblock Blip-2: Bootstrapping language-image pre-training with frozen image encoders and large language models.
\newblock {\em Pages  19730--19742 of:} {\em International conference on machine learning}.
\newblock PMLR.

\bibitem[\protect\citename{Li {\em et~al.}, }2024b]{li2024autoregressive}
Li, Tianhong, Tian, Yonglong, Li, He, Deng, Mingyang, \& He, Kaiming. 2024b.
\newblock Autoregressive image generation without vector quantization.
\newblock {\em Advances in Neural Information Processing Systems}, {\bf 37}, 56424--56445.

\bibitem[\protect\citename{Li {\em et~al.}, }2025]{li2025diffueraserdiffusionmodelvideo}
Li, Xiaowen, Xue, Haolan, Ren, Peiran, \& Bo, Liefeng. 2025.
\newblock {\em DiffuEraser: A Diffusion Model for Video Inpainting}.

\bibitem[\protect\citename{Liew {\em et~al.}, }2023]{liew2023magicedit}
Liew, Jun~Hao, Yan, Hanshu, Zhang, Jianfeng, Xu, Zhongcong, \& Feng, Jiashi. 2023.
\newblock Magicedit: High-fidelity and temporally coherent video editing.
\newblock {\em arXiv preprint arXiv:2308.14749}.

\bibitem[\protect\citename{Lin {\em et~al.}, }2024]{lin2024vila}
Lin, Ji, Yin, Hongxu, Ping, Wei, Molchanov, Pavlo, Shoeybi, Mohammad, \& Han, Song. 2024.
\newblock Vila: On pre-training for visual language models.
\newblock {\em Pages  26689--26699 of:} {\em Proceedings of the IEEE/CVF conference on computer vision and pattern recognition}.

\bibitem[\protect\citename{Liu {\em et~al.}, }2024a]{liu2024texoct}
Liu, Jialun, Wu, Chenming, Liu, Xinqi, Liu, Xing, Wu, Jinbo, Peng, Haotian, Zhao, Chen, Feng, Haocheng, Liu, Jingtuo, \& Ding, Errui. 2024a.
\newblock Texoct: Generating textures of 3d models with octree-based diffusion.
\newblock {\em Pages  4284--4293 of:} {\em Proceedings of the IEEE/CVF Conference on Computer Vision and Pattern Recognition}.

\bibitem[\protect\citename{Liu {\em et~al.}, }2025a]{liu2025texgarment}
Liu, Jialun, Wu, Jinbo, Gao, Xiaobo, Hu, Jiakui, Xiong, Bojun, Liu, Xing, Zhao, Chen, Pei, Hongbin, Feng, Haocheng, Li, Yingying, {\em et~al.} 2025a.
\newblock TexGarment: Consistent Garment UV Texture Generation via Efficient 3D Structure-Guided Diffusion Transformer.
\newblock {\em Pages  26566--26575 of:} {\em Proceedings of the Computer Vision and Pattern Recognition Conference}.

\bibitem[\protect\citename{Liu {\em et~al.}, }2025b]{liu2025infinitystar}
Liu, Jinlai, Han, Jian, Yan, Bin, Wu, Hui, Zhu, Fengda, Wang, Xing, Jiang, Yi, Peng, Bingyue, \& Yuan, Zehuan. 2025b.
\newblock Infinitystar: Unified spacetime autoregressive modeling for visual generation.
\newblock {\em arXiv preprint arXiv:2511.04675}.

\bibitem[\protect\citename{Liu {\em et~al.}, }2024b]{liu2024video}
Liu, Shaoteng, Zhang, Yuechen, Li, Wenbo, Lin, Zhe, \& Jia, Jiaya. 2024b.
\newblock Video-p2p: Video editing with cross-attention control.
\newblock {\em Pages  8599--8608 of:} {\em Proceedings of the IEEE/CVF Conference on Computer Vision and Pattern Recognition}.

\bibitem[\protect\citename{Lu {\em et~al.}, }2022]{lu2022dpm}
Lu, Cheng, Zhou, Yuhao, Bao, Fan, Chen, Jianfei, Li, Chongxuan, \& Zhu, Jun. 2022.
\newblock Dpm-solver: A fast ode solver for diffusion probabilistic model sampling in around 10 steps.
\newblock {\em Advances in neural information processing systems}, {\bf 35}, 5775--5787.

\bibitem[\protect\citename{Medeiros, }2025]{langsam}
Medeiros, Luca. 2025.
\newblock {\em lang-segment-anything}.
\newblock GitHub repository.

\bibitem[\protect\citename{Mou {\em et~al.}, }2024]{mou2024t2i}
Mou, Chong, Wang, Xintao, Xie, Liangbin, Wu, Yanze, Zhang, Jian, Qi, Zhongang, \& Shan, Ying. 2024.
\newblock T2i-adapter: Learning adapters to dig out more controllable ability for text-to-image diffusion models.
\newblock {\em Pages  4296--4304 of:} {\em Proceedings of the AAAI conference on artificial intelligence},  vol. 38.

\bibitem[\protect\citename{Peebles \& Xie, }2023]{peebles2023scalable}
Peebles, William, \& Xie, Saining. 2023.
\newblock Scalable diffusion models with transformers.
\newblock {\em Pages  4195--4205 of:} {\em Proceedings of the IEEE/CVF international conference on computer vision}.

\bibitem[\protect\citename{Podell {\em et~al.}, }2023]{podell2023sdxl}
Podell, Dustin, English, Zion, Lacey, Kyle, Blattmann, Andreas, Dockhorn, Tim, M{\"u}ller, Jonas, Penna, Joe, \& Rombach, Robin. 2023.
\newblock Sdxl: Improving latent diffusion models for high-resolution image synthesis.
\newblock {\em arXiv preprint arXiv:2307.01952}.

\bibitem[\protect\citename{Poole {\em et~al.}, }2022]{poole2022dreamfusion}
Poole, Ben, Jain, Ajay, Barron, Jonathan~T, \& Mildenhall, Ben. 2022.
\newblock Dreamfusion: Text-to-3d using 2d diffusion.
\newblock {\em arXiv preprint arXiv:2209.14988}.

\bibitem[\protect\citename{Radford {\em et~al.}, }2021]{radford2021learning}
Radford, Alec, Kim, Jong~Wook, Hallacy, Chris, Ramesh, Aditya, Goh, Gabriel, Agarwal, Sandhini, Sastry, Girish, Askell, Amanda, Mishkin, Pamela, Clark, Jack, {\em et~al.} 2021.
\newblock Learning transferable visual models from natural language supervision.
\newblock {\em Pages  8748--8763 of:} {\em International conference on machine learning}.
\newblock PmLR.

\bibitem[\protect\citename{Ramesh {\em et~al.}, }2022]{ramesh2022hierarchical}
Ramesh, Aditya, Dhariwal, Prafulla, Nichol, Alex, Chu, Casey, \& Chen, Mark. 2022.
\newblock Hierarchical text-conditional image generation with clip latents.
\newblock {\em arXiv preprint arXiv:2204.06125}, {\bf 1}(2), 3.

\bibitem[\protect\citename{Ravi {\em et~al.}, }2024]{ravi2024sam2}
Ravi, Nikhila, Gabeur, Valentin, Hu, Yuan-Ting, Hu, Ronghang, Ryali, Chaitanya, Ma, Tengyu, Khedr, Haitham, R{\"a}dle, Roman, Rolland, Chloe, Gustafson, Laura, {\em et~al.} 2024.
\newblock SAM 2: Segment Anything in Images and Videos.
\newblock {\em arXiv preprint arXiv:2408.00714}.

\bibitem[\protect\citename{Ren {\em et~al.}, }2024]{ren2024grounded}
Ren, Tianhe, Liu, Shilong, Zeng, Ailing, Lin, Jing, Li, Kunchang, Cao, He, Chen, Jiayu, Huang, Xinyu, Chen, Yukang, Yan, Feng, Zeng, Zhaoyang, Zhang, Hao, Li, Feng, Yang, Jie, Li, Hongyang, Jiang, Qing, \& Zhang, Lei. 2024.
\newblock {\em Grounded SAM: Assembling Open-World Models for Diverse Visual Tasks}.

\bibitem[\protect\citename{Saharia {\em et~al.}, }2022]{saharia2022photorealistic}
Saharia, Chitwan, Chan, William, Saxena, Saurabh, Li, Lala, Whang, Jay, Denton, Emily~L, Ghasemipour, Kamyar, Gontijo~Lopes, Raphael, Karagol~Ayan, Burcu, Salimans, Tim, {\em et~al.} 2022.
\newblock Photorealistic text-to-image diffusion models with deep language understanding.
\newblock {\em Advances in neural information processing systems}, {\bf 35}, 36479--36494.

\bibitem[\protect\citename{Seedream {\em et~al.}, }2025]{seedream2025seedream}
Seedream, Team, Chen, Yunpeng, Gao, Yu, Gong, Lixue, Guo, Meng, Guo, Qiushan, Guo, Zhiyao, Hou, Xiaoxia, Huang, Weilin, Huang, Yixuan, {\em et~al.} 2025.
\newblock Seedream 4.0: Toward next-generation multimodal image generation.
\newblock {\em arXiv preprint arXiv:2509.20427}.

\bibitem[\protect\citename{Sheynin {\em et~al.}, }2024]{sheynin2024emu}
Sheynin, Shelly, Polyak, Adam, Singer, Uriel, Kirstain, Yuval, Zohar, Amit, Ashual, Oron, Parikh, Devi, \& Taigman, Yaniv. 2024.
\newblock Emu edit: Precise image editing via recognition and generation tasks.
\newblock {\em Pages  8871--8879 of:} {\em Proceedings of the IEEE/CVF Conference on Computer Vision and Pattern Recognition}.

\bibitem[\protect\citename{Shi {\em et~al.}, }2023]{shi2023zero123++}
Shi, Ruoxi, Chen, Hansheng, Zhang, Zhuoyang, Liu, Minghua, Xu, Chao, Wei, Xinyue, Chen, Linghao, Zeng, Chong, \& Su, Hao. 2023.
\newblock Zero123++: a single image to consistent multi-view diffusion base model.
\newblock {\em arXiv preprint arXiv:2310.15110}.

\bibitem[\protect\citename{Singer {\em et~al.}, }2022]{singer2022make}
Singer, Uriel, Polyak, Adam, Hayes, Thomas, Yin, Xi, An, Jie, Zhang, Songyang, Hu, Qiyuan, Yang, Harry, Ashual, Oron, Gafni, Oran, {\em et~al.} 2022.
\newblock Make-A-Video: Text-to-Video Generation without Text-Video Data.
\newblock {\em arXiv preprint arXiv:2209.14792}.

\bibitem[\protect\citename{Tan {\em et~al.}, }2025]{tan2025ominicontrol}
Tan, Zhenxiong, Liu, Songhua, Yang, Xingyi, Xue, Qiaochu, \& Wang, Xinchao. 2025.
\newblock Ominicontrol: Minimal and universal control for diffusion transformer.
\newblock {\em Pages  14940--14950 of:} {\em Proceedings of the IEEE/CVF International Conference on Computer Vision}.

\bibitem[\protect\citename{Tan {\em et~al.}, }2024]{tan2024vidgen}
Tan, Zhiyu, Yang, Xiaomeng, Qin, Luozheng, \& Li, Hao. 2024.
\newblock Vidgen-1m: A large-scale dataset for text-to-video generation.
\newblock {\em arXiv preprint arXiv:2408.02629}.

\bibitem[\protect\citename{Team {\em et~al.}, }2023]{gemini}
Team, Gemini, Anil, Rohan, Borgeaud, Sebastian, Alayrac, Jean-Baptiste, Yu, Jiahui, Soricut, Radu, Schalkwyk, Johan, Dai, Andrew~M, Hauth, Anja, Millican, Katie, {\em et~al.} 2023.
\newblock Gemini: a family of highly capable multimodal models.
\newblock {\em arXiv preprint arXiv:2312.11805}.

\bibitem[\protect\citename{Tian {\em et~al.}, }2024]{tian2024visual}
Tian, Keyu, Jiang, Yi, Yuan, Zehuan, Peng, Bingyue, \& Wang, Liwei. 2024.
\newblock Visual autoregressive modeling: Scalable image generation via next-scale prediction.
\newblock {\em Advances in neural information processing systems}, {\bf 37}, 84839--84865.

\bibitem[\protect\citename{Tu {\em et~al.}, }2025]{tu2025videoanydoor}
Tu, Yuanpeng, Luo, Hao, Chen, Xi, Ji, Sihui, Bai, Xiang, \& Zhao, Hengshuang. 2025.
\newblock Videoanydoor: High-fidelity video object insertion with precise motion control.
\newblock {\em Pages  1--11 of:} {\em Proceedings of the Special Interest Group on Computer Graphics and Interactive Techniques Conference Conference Papers}.

\bibitem[\protect\citename{Villegas {\em et~al.}, }2022]{villegas2022phenaki}
Villegas, Ruben, Babaeizadeh, Mohammad, Kindermans, Pieter-Jan, Moraldo, Hernan, Zhang, Han, Saffar, Mohammad~Taghi, Castro, Santiago, Kunze, Julius, \& Erhan, Dumitru. 2022.
\newblock Phenaki: Variable length video generation from open domain textual description.
\newblock {\em arXiv preprint arXiv:2210.02399}.

\bibitem[\protect\citename{Wan {\em et~al.}, }2025]{wan2025wan}
Wan, Team, Wang, Ang, Ai, Baole, Wen, Bin, Mao, Chaojie, Xie, Chen-Wei, Chen, Di, Yu, Feiwu, Zhao, Haiming, Yang, Jianxiao, {\em et~al.} 2025.
\newblock Wan: Open and advanced large-scale video generative models.
\newblock {\em arXiv preprint arXiv:2503.20314}.

\bibitem[\protect\citename{Wang {\em et~al.}, }2024a]{wang2024qwen2}
Wang, Peng, Bai, Shuai, Tan, Sinan, Wang, Shijie, Fan, Zhihao, Bai, Jinze, Chen, Keqin, Liu, Xuejing, Wang, Jialin, Ge, Wenbin, {\em et~al.} 2024a.
\newblock Qwen2-vl: Enhancing vision-language model's perception of the world at any resolution.
\newblock {\em arXiv preprint arXiv:2409.12191}.

\bibitem[\protect\citename{Wang {\em et~al.}, }2025]{wang2025koala}
Wang, Qiuheng, Shi, Yukai, Ou, Jiarong, Chen, Rui, Lin, Ke, Wang, Jiahao, Jiang, Boyuan, Yang, Haotian, Zheng, Mingwu, Tao, Xin, {\em et~al.} 2025.
\newblock Koala-36m: A large-scale video dataset improving consistency between fine-grained conditions and video content.
\newblock {\em Pages  8428--8437 of:} {\em Proceedings of the Computer Vision and Pattern Recognition Conference}.

\bibitem[\protect\citename{Wang {\em et~al.}, }2024b]{wang2024replace}
Wang, Xiang, Zhang, Shiwei, Qiu, Haonan, Chu, Ruihang, Li, Zekun, Zhang, Yingya, Gao, Changxin, Wang, Yuehuan, Shen, Chunhua, \& Sang, Nong. 2024b.
\newblock Replace anyone in videos.
\newblock {\em arXiv preprint arXiv:2409.19911}.

\bibitem[\protect\citename{Wang {\em et~al.}, }2024c]{wang2024causal}
Wang, Yuqing, Li, Xiangxian, Liu, Yannan, Cao, Xiao, Meng, Xiangxu, \& Meng, Lei. 2024c.
\newblock Causal inference for out-of-distribution recognition via sample balancing.
\newblock {\em CAAI Transactions on Intelligence Technology}, {\bf 9}(5), 1172--1184.

\bibitem[\protect\citename{Wang {\em et~al.}, }2024d]{wang2024motionctrl}
Wang, Zhouxia, Yuan, Ziyang, Wang, Xintao, Li, Yaowei, Chen, Tianshui, Xia, Menghan, Luo, Ping, \& Shan, Ying. 2024d.
\newblock Motionctrl: A unified and flexible motion controller for video generation.
\newblock {\em In:} {\em ACM SIGGRAPH 2024 Conference Papers}.

\bibitem[\protect\citename{Wei {\em et~al.}, }2025]{wei2025univideo}
Wei, Cong, Liu, Quande, Ye, Zixuan, Wang, Qiulin, Wang, Xintao, Wan, Pengfei, Gai, Kun, \& Chen, Wenhu. 2025.
\newblock Univideo: Unified understanding, generation, and editing for videos.
\newblock {\em arXiv preprint arXiv:2510.08377}.

\bibitem[\protect\citename{Wu {\em et~al.}, }2025]{wu2025janus}
Wu, Chengyue, Chen, Xiaokang, Wu, Zhiyu, Ma, Yiyang, Liu, Xingchao, Pan, Zizheng, Liu, Wen, Xie, Zhenda, Yu, Xingkai, Ruan, Chong, {\em et~al.} 2025.
\newblock Janus: Decoupling visual encoding for unified multimodal understanding and generation.
\newblock {\em Pages  12966--12977 of:} {\em Proceedings of the Computer Vision and Pattern Recognition Conference}.

\bibitem[\protect\citename{Xiao {\em et~al.}, }2025]{xiao2025omnigen}
Xiao, Shitao, Wang, Yueze, Zhou, Junjie, Yuan, Huaying, Xing, Xingrun, Yan, Ruiran, Li, Chaofan, Wang, Shuting, Huang, Tiejun, \& Liu, Zheng. 2025.
\newblock Omnigen: Unified image generation.
\newblock {\em Pages  13294--13304 of:} {\em Proceedings of the Computer Vision and Pattern Recognition Conference}.

\bibitem[\protect\citename{Xiong {\em et~al.}, }2025]{xiong2025texgaussian}
Xiong, Bojun, Liu, Jialun, Hu, Jiakui, Wu, Chenming, Wu, Jinbo, Liu, Xing, Zhao, Chen, Ding, Errui, \& Lian, Zhouhui. 2025.
\newblock Texgaussian: Generating high-quality pbr material via octree-based 3d gaussian splatting.
\newblock {\em Pages  551--561 of:} {\em Proceedings of the Computer Vision and Pattern Recognition Conference}.

\bibitem[\protect\citename{Yang {\em et~al.}, }2025]{yang2025not}
Yang, Liying, Liu, Chen, Zhu, Zhenwei, Liu, Ajian, Ma, Hui, Nong, Jian, \& Liang, Yanyan. 2025 (October).
\newblock Not All Frame Features Are Equal: Video-to-4D Generation via Decoupling Dynamic-Static Features.
\newblock {\em Pages  7494--7504 of:} {\em Proceedings of the IEEE/CVF International Conference on Computer Vision (ICCV)}.

\bibitem[\protect\citename{Yu {\em et~al.}, }2023]{yu2023language}
Yu, Lijun, Lezama, Jos{\'e}, Gundavarapu, Nitesh~B, Versari, Luca, Sohn, Kihyuk, Minnen, David, Cheng, Yong, Birodkar, Vighnesh, Gupta, Agrim, Gu, Xiuye, {\em et~al.} 2023.
\newblock Language Model Beats Diffusion--Tokenizer is Key to Visual Generation.
\newblock {\em arXiv preprint arXiv:2310.05737}.

\bibitem[\protect\citename{Yuan {\em et~al.}, }2025]{yuan2025opens2v}
Yuan, Shenghai, He, Xianyi, Deng, Yufan, Ye, Yang, Huang, Jinfa, Lin, Bin, Luo, Jiebo, \& Yuan, Li. 2025.
\newblock Opens2v-nexus: A detailed benchmark and million-scale dataset for subject-to-video generation.
\newblock {\em arXiv preprint arXiv:2505.20292}.

\bibitem[\protect\citename{Zhang {\em et~al.}, }2023]{zhang2023adding}
Zhang, Lvmin, Rao, Anyi, \& Agrawala, Maneesh. 2023.
\newblock Adding conditional control to text-to-image diffusion models.
\newblock {\em In:} {\em Proceedings of the IEEE/CVF international conference on computer vision}.

\bibitem[\protect\citename{Zhang {\em et~al.}, }2026]{zhang2026omnitransfer}
Zhang, Pengze, Wu, Yanze, Li, Mengtian, Bai, Xu, Zhao, Songtao, Ye, Fulong, Mou, Chong, Li, Xinghui, Chen, Zhuowei, He, Qian, {\em et~al.} 2026.
\newblock OmniTransfer: All-in-one Framework for Spatio-temporal Video Transfer.
\newblock {\em arXiv preprint arXiv:2601.14250}.

\bibitem[\protect\citename{Zheng {\em et~al.}, }2024]{zheng2024open}
Zheng, Zangwei, Peng, Xiangyu, Yang, Tianji, Shen, Chenhui, Li, Shenggui, Liu, Hongxin, Zhou, Yukun, Li, Tianyi, \& You, Yang. 2024.
\newblock Open-sora: Democratizing efficient video production for all.
\newblock {\em arXiv preprint arXiv:2412.20404}.

\bibitem[\protect\citename{Zi {\em et~al.}, }2025]{minimaxremover}
Zi, Bojia, Peng, Weixuan, Qi, Xianbiao, Wang, Jianan, Zhao, Shihao, Xiao, Rong, \& Wong, Kam-Fai. 2025.
\newblock {\em MiniMax-Remover: Taming Bad Noise Helps Video Object Removal}.

\end{thebibliography}
\bibliographystyle{authordate1}

\end{document}